\documentclass[10pt,conference]{IEEEtran}
\IEEEoverridecommandlockouts
\usepackage{amsmath}
\usepackage{graphicx}
\usepackage{cite}
\usepackage{amsmath,amssymb,amsfonts}
\usepackage{subcaption}
\usepackage{algorithm}
\usepackage{algpseudocode}
\usepackage{textcomp}
\usepackage{placeins}
\usepackage{xcolor}
\usepackage{float}
\usepackage{mathrsfs}
\usepackage{titlesec}
\usepackage{subcaption}
\usepackage{caption}
\usepackage{layout}
\titlespacing{\section}{1.5pt}{1ex}{1ex} 
\titlespacing{\subsection}{1pt}{1ex}{1ex} 
\ifCLASSINFOpdf
\else
\fi
\begin{document}
\title{Joint UAV-UGV Positioning and Trajectory Planning via Meta A3C for Reliable Emergency Communications}
\author{
    Ndagijimana~Cyprien\textsuperscript{*}, 
    Mehdi~Sookhak\textsuperscript{*}, 
    Hosein~Zarini\textsuperscript{*},
    Chandra~N~Sekharan\textsuperscript{*}, 
    and Mohammed~Atiquzzaman\textsuperscript{\dag}
    
    \\
    \textsuperscript{*}Department of Computer Science, Texas A\&M University-Corpus Christi, Tx, USA \\
    \textsuperscript{\dag} School of Computer Science, University of Oklahoma, Norman, OK USA\\
    Emails: cndagijimana@islander.tamucc.edu; (mehdi.sookhak, chandra.sekharan)@tamucc.edu); atiq@ou.edu
}

\markboth{Journal of \LaTeX\ Class Files,~Vol.~14, No.~8, August~2015}%
{Shell \MakeLowercase{\textit{et al.}}: Bare Demo of IEEEtran.cls for IEEE Journals}
\maketitle
\begin{abstract}
Joint deployment of unmanned aerial vehicles (UAVs) and unmanned ground vehicles (UGVs) has been shown to be an effective method to establish communications in areas affected by disasters. However, ensuring good Quality of Services (QoS) while using as few UAVs as possible also requires optimal positioning and trajectory planning for UAVs and UGVs. This paper proposes a joint UAV-UGV-based positioning and trajectory planning framework for UAVs and UGVs deployment that guarantees optimal QoS for ground users. To model the UGVs' mobility, we introduce a road graph, which directs their movement along valid road segments and adheres to the road network constraints. To solve the sum rate optimization
problem, we reformulate the problem as a Markov Decision Process (MDP) and propose a novel asynchronous Advantage Actor Critic (A3C) incorporated with meta-learning for rapid adaptation to new environments and dynamic conditions. Numerical results demonstrate that our proposed Meta-A3C approach outperforms A3C and DDPG, delivering 13.1\% higher throughput and 49\% faster execution while meeting the QoS requirements.
\end{abstract}
\IEEEpeerreviewmaketitle
\section{Introduction}
Unmanned Ground Vehicles (UGVs) have been proposed as a promising solution to provide backhaul links to Unmanned Aerial Vehicles (UAVs) in case terrestrial base stations (BSs) are compromised \cite{b1,b2}. In these recovery scenarios, UGVs are deployed to provide stable, high-capacity links to the UAVs acting as flying base stations and also to provide mobile wireless coverage to ground users. Owing to their enhanced payload capacity and extended energy supplies, UGVs enable the mounting of advanced communication equipment while maintaining continuous ground operations, thereby enabling fast restoration of critical network infrastructure \cite{b3, b4}. The UGV-UAVs combined framework is vital for rapidly deployable and resilient networks that can be adaptive to dynamic emergency scenarios and QoS requirements.

However, the energy efficiency (EE), optimal positioning, mobility, and trajectory planning remain significant challenges for UGV-UAV wireless networks. To tackle these issues, substantial research has concentrated on optimal positioning \cite{b5,b24,b25,b27,b28}, trajectory planning \cite{b6,b7}, and resource allocation strategies \cite{b9}. An efficient 3D positioning algorithm was proposed to minimize the number of UAVs required while optimizing their deployment positions \cite{b10}. In \cite{b11}, a deep reinforcement learning (DRL) approach was proposed to jointly optimize the 3D trajectory of UAVs and minimize UAV propulsion energy. Addressing user-side power constraints, the authors in \cite{b12} proposed a safe Deep Q-Network (DQN)-based UAV trajectory optimization framework aimed at maximizing uplink throughput while ensuring energy efficiency. Moreover, authors in \cite{b13} proposed a deep supervised learning approach for joint optimization of UAV caching and trajectory planning, and authors in \cite{b14} proposed a semidefinite relaxation-based method for 3D trajectory optimization. 

Despite these advancements, the integration of UGVs to improve the backhaul connectivity of UAVs has not been thoroughly investigated. Particularly, the joint optimization of positioning and trajectory for both UAVs and UGVs remains an underexplored area. This work aims to develop an intelligent framework for joint optimal positioning and trajectory design of UAVs and UGVs to maximize network throughput, while satisfying QoS requirements for users in disaster-affected areas. Although prior studies have addressed UAV positioning and trajectory planning, they typically do not consider the joint operation of UAV-UGV systems in dynamic environments that involve UAV-to-UGV-to-user communication links \cite{b15,b16}. 

The key distinction of our work lies in the simultaneous optimization of UAV and UGV positioning and trajectories, and the novel integration of meta-learning with the Asynchronous Advantage Actor-Critic (A3C) algorithm. This combination enables rapid adaptation to environmental dynamics. Specifically, our proposed unified communication framework among UAVs, UGVs, and users, augmented with trajectory optimization, is designed to deliver optimal system performance and improved QoS in real-time, dynamically changing environments. \textit{To the best of our knowledge, this is the first work to present an integrated framework that jointly optimizes the positioning and trajectories of both UAVs and UGVs in a dynamic communication environment, with the goal of maximizing network performance and user QoS.} The contributions of this paper are as follows:

\begin{enumerate}
    \item[\Large\textbullet] We jointly formulated an optimization problem to maximize the sum rate by optimizing UGV-UAV and UAV-user associations, while ensuring constraints on distance, altitude, speed, UAV separation, and UGV movement within the defined road network.
    \item[\Large\textbullet] Due to the non-convexity and the complexity of our optimization problem, we carefully reformulate the problem into a Markov Decision Process (MDP) to enable dynamic modeling of the system's behavior.
    \item[\Large\textbullet] We then introduce an A3C-based framework for UAV and UGV positioning and trajectory planning, designed to ensure that the QoS requirements of ground users are consistently met.
    \item[\Large\textbullet] To enable real-time deployment in emergency response scenarios, we integrate a meta-learning approach with the A3C model, facilitating rapid adaptation to dynamic channel conditions and evolving user demands.
    \item[\Large\textbullet] Finally, simulated results demonstrated that the proposed Meta-A3C approach achieves 13.1\% higher throughput than A3C and 30.1\% than DDPG methods with low complexity.
\end{enumerate}

The remainder of this paper is arranged as follows: Section II provides the system model and problem formulation, while Section III presents the Meta-A3C approach. Evaluation results are discussed in Section IV, and the conclusion is given in Section V.
\section{System Model and Problem Formulation}

\subsection{Mathematical Modeling of UGV Trajectories}
We consider multi-UGV-UAV cooperative networks deployed in an emergency management scenario, as illustrated in Figure 1. The system consists of $M$ UGVs denoted by a set $m \in \mathcal{M} = \{1, 2, ..., M\}$ which establish a backhaul connectivity to $U$ UAVs, denoted by a set $\mathcal{U} = \{1, 2, \dots, U\}$. The total operational time $T$ is divided into $N$ discrete time slots, each of duration $\Delta$ seconds, so that $T = N\Delta$. We consider geographical features of the road network modeled as a graph representation denoted as $ G = [\mathcal{V}, \mathcal{E}]$. The nodes $\mathcal{V} = \{\mathcal{V}_1,\mathcal{V}_2,...,\mathcal{V}_{\mathcal{Q}}\}$ represents intersections where $\mathcal{Q}$ is the number of intersections on the road network. The edges $ \mathcal{E} = \{\mathcal{E}_{ij} = (\mathcal{V}_i, \mathcal{V}_j)\}, \forall i,j \in \mathcal{\mathcal{Q}},$ denotes the set of road segments. For each time step $n \in \mathcal{N} = \{1, 2, ..., N\} $, the position of $m^{th}$ UGV is defined as $\boldsymbol p_m [n] = [x_m[n], y_m[n],0] \in \mathcal{E}$ with initial position $ \boldsymbol {p}_m[0] = [(x_m[0], y_m[0], 0)]$ for $\mathcal{E}_{ij} \in \mathcal{E}$. The $m^{th}$ UGV trajectory for each time slot $n$ is given by $ \boldsymbol {p}_m = [\boldsymbol{p}_m[1], \boldsymbol{p}_m[2], ..., \boldsymbol{p}_m[n],..,\boldsymbol{p}_m[N]]$. In addition, the $m^{th}$ UGV's velocity along $\mathcal{E}_{ij} $ edge is constrained by $0 \leq v_{m}[n] \leq \min(v_{ij}^{\max}, V^{\max}_m)$ such that:
\begin{equation}
v_{m}[n]=\frac{\left\Vert \boldsymbol{p}_m [n] - \boldsymbol{p}_m[n-1] \right\Vert}{\Delta} \leq V^{\max}_m,
\end{equation}
where $0 \leq v_{m}[n] \leq \min(v_{ij}^{\max}, V^{\max}_m), \forall i,j \in \{1, 2, ..., \mathcal{\mathcal{Q}}\}$, $v_{ij}^{\max}$ is the road segment speed limit and $V^{\max}_m$ is the maximum UGV's velocity. We impose the condition  $\boldsymbol{p}_m[N] = \boldsymbol{p}_m[0] $, requiring UGV to return to its initial position after completing the mission tasks. 
\begin{figure}[t]
  \centering
  \includegraphics[width=0.45\textwidth]{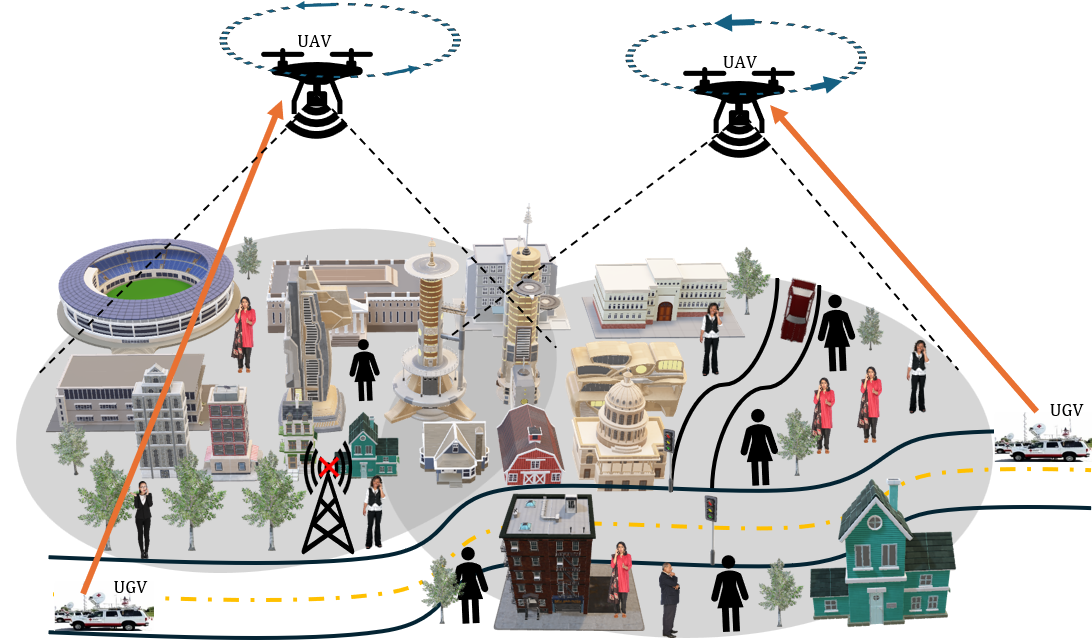}
  \caption{UAV-Assisted Wireless Networks with UGV in an Emergency Situation.}
  \label{fig:my-figure}
  \vspace{-1.5em}
\end{figure}
\subsection{Mathematical Modeling of UAV Trajectories}
Considering a system where a set of UAVs provides wireless connectivity to $K$ ground users, denoted by a set $\mathcal{K} = \{1, 2, \dots, K\}$.  We define the position of the $u^{th}$ UAV and $k^{th}$ user for each time slot $ n $ as $\boldsymbol{q}_u[n] = [x_u[n], y_u[n], z_u[n]]$ and $\boldsymbol{p}_k[n] = [x_k[n], y_k[n], 0]$, respectively. Moreover, the initial position $u^{th}$ UAV is $ \boldsymbol{q}_u[0] = [x_u[0], y_u[0], z_u[0]] $ and trajectory in each time slot $n$ is $\boldsymbol{q}_u = [\boldsymbol{q}_u[1], \boldsymbol{q}_u[2], ..., \boldsymbol{q}_u[n],..,\boldsymbol{q}_u[N]] , \forall n \in \mathcal{N}$. We consider that the $u^{th}$ UAV moves with a velocity ${v}_u[n]$, constrained by its maximum velocity such that: 
\begin{equation}
{v}_u[n] =\frac{\left\Vert \boldsymbol{q}_u [n] - \boldsymbol{q}_u[n-1] \right\Vert}{\Delta} \leq V^{\max}_u, \forall n \in \mathcal{N}
\end{equation}
where $\|\cdot\|$ denotes the Euclidean norm. Under the condition $\boldsymbol{q}_u [0] = \boldsymbol{q}_u [N]$, each UAV must return to its initial position once tasks are completed.
\subsection{G2A and A2G Wireless Channel Models}
The wireless channels between $m^{th}$ UGV and $u^{th}$ UAV and between $u^{th}$ UAV and $k^{th}$ ground user are predominantly influenced by the transmission distances, propagation environments, and elevation angles. We model both links for G2A using a probabilistic path loss approach, accounting for LoS and NLoS components as expressed by \cite{b17}: 
\begin{equation}
\Gamma_{m,u}^{\text{LoS}}[n] = 20 \log_{10} \left( \frac{4\pi f_c d_{m,u}[n]}{c} \right) + \beta^{\text{LoS}}
\end{equation}
\begin{equation}
\mathcal{F}_{m,u}^{\text{NLoS}}[n] = 20 \log_{10} \left( \frac{4\pi f_c d_{m,u}[n]}{c} \right) + \beta^{\text{NLoS}}
\end{equation}
where $f_c$ is the carrier frequency, $c=3\cdot10^8 m /s$ is the speed of light and $d_{m,u}[n] = \left\Vert \boldsymbol{q}_u [n] - \boldsymbol{p}_m[n] \right\Vert$ is the G2A distance from $m^{th}$ UGV to $u^{th}$ UAV. Accordingly, the A2G links are modeled using the same method. The probability of establishing the LoS link from $m^{th}$ UGV to $u^{th}$ UAV as well as $u^{th}$ UAV to $k^{th}$ user can be given by:
\begin{equation}
  P^{\text{los}}_{m,u}[n] = \frac{1}{1 + \eta_1 \exp\left(-\eta_2 \left[\psi_{m,u}[n] - \eta_1 \right]\right)},
\end{equation}
\begin{equation}
  P^{\text{los}}_{u,k}[n] = \frac{1}{1 + \beta_1 \exp\left(-\beta_2 \left[\psi_{u,k}[n] - \beta_1 \right]\right)},
\end{equation}
where $\eta_1$ and $\eta_2$ are environment dependent variables,  $\psi_{u,k}[n] = \frac{180}{\pi} \sin^{-1} \left( \frac{z_u[n]}{d_{u,k}[n]} \right)$ is the elevation angle between $k^{th}$ user and $u^{th}$ UAV, where $d_{u,k}[n] = \left\Vert \boldsymbol{q}_u [n] - \boldsymbol{p}_k[n] \right\Vert$ is the A2G distance. The probability of path losses for G2A and A2G links is given by: 
\begin{equation}
\begin{aligned}
L_{m,u}[n]  &= \Gamma^{\text{loS}}_{m,u}[n] P^{\text{loS}}_{m,u}[n] + \mathcal{F}_{m,u}^{\text{NLoS}}[n] P^{\text{NloS}}_{m,u}[n]
\end{aligned}
\end{equation}
\begin{equation}
\begin{aligned}
L_{u,k}[n]  &= \Gamma^{\text{loS}}_{u,k}[n] P^{\text{loS}}_{u,k}[n] + \mathcal{F}_{u,k}^{\text{NLoS}}[n] P^{\text{NloS}}_{u,k}[n]
\end{aligned}
\end{equation}
Therefore, the received signal-to-interference plus noise ratio (SINR) at the $k$-th user from $u$-th UAV is given by: 
\begin{equation}
\text{SINR}_{u,k} = \frac{P_u[n]  10^{-L_{u,k}[n]/10}}{\sum\limits_{\, u' \ne u} P_{u'}[n] 10^{-L_{u',k}[n]/10} + \sigma_k^2}, \hspace{0.1cm} \forall u' \in \mathcal{U},
\end{equation}
where $0 \leq  P_{u}[n] \leq P^{\max}_{u}$, $P_{u}[n]$ is the transmit power per UAV and \( \sigma_{k}^2 \) is the noise power at the \( k \) user, \( \mathcal{B} \) denotes the available bandwidth. Then the achievable data rate at the $k$-th user from the $u$-th UAV is calculated as follows: 
\begin{equation}
R_{u,k}[n] =    \alpha_{u,k}[n] \mathcal{B}\log_2 \left( 1 +\text{SINR}_{u,k}  \right), \hspace{0.1cm} \forall k \in \mathcal{K},
\end{equation}
where $\alpha_{u,k}[n]$ is the user association binary variable, $ \alpha_{u,k}[n] = 1$ if $k$-th  user is associated to $u^{th}$ UAV, and $0$ otherwise.
Moreover, we define a binary variable $x_{m,u}[n]$, where $x_{m,u}[n] = 1$ when $m^{th}$ UGV is associated with $u^{th}$ UAV, and $0$ otherwise. 
\subsection{ Problem Formulation}
We aim to maximize the sum rate for our system by jointly optimizing the UAV and UGV positioning and trajectory, ensuring QoS requirements for both UAVs and users, as formulated below: 
\begin{equation}
\begin{aligned}
& \max_{\{\mathbf{q}\}, \{\mathbf{p}\}, \{\mathbf{x}\},\{\mathbf{\alpha}\}} R_{\text{sum}} =  \sum_{u=1}^{U} \sum_{k=1}^{K}   R_{u,k}[n] \\
& \text{Subject to:} \\
& \hspace{1cm} C_1: \text{SINR}_{m,u} \geq \text{SINR}_u^{min}, \hspace{0.1cm} \forall u \in \mathcal{U},\\
& \hspace{1cm} C_2: R_{k}[n] \geq R_k^{\text{min}},  \hspace{0.1cm} \forall k \in \mathcal{K},\; \forall n \in \mathcal{N},  \\
& \hspace{1cm} C_3: {v}_u[n] =\frac{\left\Vert \boldsymbol{q}_u [n] - \boldsymbol{q}_u[n-1] \right\Vert}{\Delta} \leq V^{\max}_u, \forall u \in \mathcal{U}\; \\
& \hspace{1cm} C_4: \mathbf{q}_u[N] = \mathbf{q}_u[0], \hspace{0.1cm} \forall u \in \mathcal{U},\; \forall n \in \mathcal{N},\\
& \hspace{1cm} C_{5}: \|\mathbf{q}_u[n] - \mathbf{q}_{u'}[n]\| \geq d_{\text{safe}},  \forall u \neq u'\in \mathcal{U},\; \forall n \in \mathcal{N} \\
& \hspace{1cm} C_6: v_{m}[n]=\frac{\left\Vert \boldsymbol{p}_m [n] - \boldsymbol{p}_m[n-1] \right\Vert}{\Delta} \leq V^{\max}_m, \\
& \hspace{1cm} C_7: \mathbf{p}_m[n] \in G,  \hspace{0.1cm} \forall m \in \mathcal{M},\; \forall n \in \mathcal{N}, \\
& \hspace{1cm} C_{8}: \mathbf{p}_m[N] = \mathbf{p}_m(0), \hspace{0.1cm} \forall m \in \mathcal{M}, \forall n \in \mathcal{N}, \\
& \hspace{1cm} C_{9}: \sum_{u \in \mathcal{U}} \alpha_{u,k}[n] \leq 1, \hspace{0.1cm} \forall k \in \mathcal{K}, \forall u \in \mathcal{U},\; \forall n \in \mathcal{N}, \\
& \hspace{1cm} C_{10}: \sum_{u \in \mathcal{U}} x_{m,u}[n] \leq 1, \hspace{0.1cm} \forall m \in \mathcal{M}, \forall u \in \mathcal{U},\; \forall n \in \mathcal{N}, \\
& \hspace{1cm} C_{11}: x_{m,u}[n] \in \{0, 1\},  \alpha_{u,k}[n] \in \{0, 1\}, \hspace{0.1cm} \forall n \in \mathcal{N}. \\
\end{aligned}
\end{equation}
In above optimization problem, consider $\mathbf{q}= \{\mathbf{q}_{u}[n], \forall u, n\} $, $\mathbf{p}= \{\mathbf{p}_{m}[n], \forall m, n\} $, $\mathbf{x}= \{\mathbf{x}_{m,u}[n], \forall m, u, n\} $, and $\mathbf{\alpha}= \{\mathbf{\alpha}_{u,k}[n], \forall k, u, n\} $ as decision variable under consideration of constraints $C_1$ to $C_{11}$. Specifically, $C_1$, and $C_2$ ensure a minimum  threshold $\gamma_u^{min}$ and QoS for both $u^{th}$ UAV and $k^{th}$ user, respectively. To ensure sufficient QoS at the $u^{th}$ UAV, the backhaul link between the $u^{th}$ UAV and $m^{th}$ UGV must satisfy a minimum SNR threshold $\text{SINR}_{m,u}[n]=\frac{x_{m,u}[n]P_m[n] 10^{-L_{m,u}[n]/10}}{ \sigma^2}$, where $ P_m[n] $ is the transmit power of the $m^{th}$ UGV. $C_3$ restricts the $u^{th}$ UAV to fly with maximum speed of $v^{\max}_u$, while forcing it to return to its initial position in $C_4$. Constraint $C_5$ maintains a safe distance $d_{\text{safe}}$ separating two UAVs to avoid collision, and $C_6$ ensures that each UGV maintains the maximum speed allowable $V^{\max}_m$. Then, $C_7$ forces $m^{th}$ UGV to stay on a predefined path modeled with graph $G$, and maintains road-specific speed limits. Similarly, $C_{8}$ requires $m^{th}$ UGV to return to its initial position after task completion. $C_{9}$ ensures that each user may receive service from at most one UAV at time steps $n$. Furthermore, $C_{10}$ ensure that each $m^{th}$ UGV is associated with at most one UAV While $C_{11}$ enforces binary variables $\alpha_{uk}[n]$ and $x_{m,u}[n]$ $\in \{0, 1\}$ ($0=$ no link; $1=$ active link).
\section{Proposed META-A3C Approach}
\subsection{Markov Decision Process (MDP) Reformulation}
We model the joint UAV \& UGV positioning and trajectory planning as a MDP defined by a tuple ($\mathcal{S}$, $\mathcal{A}$, $\mathcal{P}$, $\mathcal{R}$, $\mathcal{\gamma}$, $\mathcal{H}$ ), where $\mathcal{S}$ is the state space, $\mathcal{A}$ is the action space, $\mathcal{P}$ is the transition probability, $\mathcal{R}$ is the reward, $\mathcal{\gamma}$ is the discount factor, and $\mathcal{H}$ is the time horizon.

\textbf{State Space $\mathcal{S} = \{ s[n] \mid n \in \mathcal{N} \}$}: At the time step $n$, the state $s[n]$ of each UAV and UGV includes position $\boldsymbol{q}_u[n] = [x_u[n], y_u[n], z_u[n]] $ of $u^{th}$ UAV, location of $m^{th}$ UGV $\boldsymbol p_m [n] = [x_m[n], y_m[n],0]$, coordinates of $k^{th}$ user $\boldsymbol{p}_k[n] = [x_k[n], y_k[n], 0]$, and $\text{SINR}_{u,k}[n]$ consisting of channel information. 

\textbf{Action Space $\mathcal{A}$}: At each time step $n$, each agent takes action $\mathcal{A} = \{ a[n] \mid n \in \mathcal{N} \}$ based on state space. This consists of continuous movement of UGVs and UAVs in a specific direction $ \boldsymbol {p}_m = [\boldsymbol{p}_m[1], \boldsymbol{p}_m[2], ..., \boldsymbol{p}_m[n],..,\boldsymbol{p}_m[N]]$, $\boldsymbol{q}_u = [\boldsymbol{q}_u[1], \boldsymbol{q}_u[2], ..., \boldsymbol{q}_u[n],..,\boldsymbol{q}_u[N]], \forall n \in \mathcal{N}$, decision variables 
 $\mathbf{x}= \{\mathbf{x}_{mu}[n], \forall m, u, n\} $ and $\mathbf{\alpha}= \{\mathbf{\alpha}_{uk}[n], \forall k, u, n\} $. 
\textbf{State Transition Function}: We denote $P(s_{n+1} \mid s_n, a_n) $ to represent the probability of moving an agent (i.e., UAV or UGV) from state $s_n$ to $s_{n+1}$ after implementing action $a_n$.

\textbf{Policy $\pi(a \mid s) = P(a \mid s )$}: We define the policy function $\pi$ as the decision strategy of mapping each state $ s \in \mathcal{S} $ to a probability distribution over the set of possible actions $ a \in \mathcal{A} $.
\textbf{Reward Function}: Considering our optimization problem, the reward $r[n]=\mathcal{R}(s_n,a_n)$ in one time slot $n$ aims to maximize the $k$ user data rate while ensuring QoS is met, as formulated below: 
 \begin{equation}
r(s_n,a_n) =  \sum_{k=1}^{K}  R_{k}[n] -  w\sum_{k=1}^{K}   \Xi_k ,
 \end{equation}
where $w$ is the weight to balance the reward and penalty, $\Xi_k$ is a continuous penalty, where $ \Xi_k =0 $ if $R_{k}[n] \geq R_k^{\text{min}}$, and 0 otherwise. 
\subsection{A3C Algorithm}
The A3C framework enables efficient state space exploration through multiple parallel actor-learners with inherent stability via the policy gradient update that enhances convergence and supports smooth trajectory planning via continuous action state compatibility. Usually, the actor network, parameterized by \( \varphi_a \), chooses the action to be taken by following a policy \( \pi(a|s; \varphi_a) = \mathcal{P} (a|s; \varphi_a) \). The update of these parameters is carried out using policy gradient approaches. 

On the other hand, the critic network parameterized by \( \varphi_c \) estimates the state-value function \( V^{\pi}(s_n; \varphi_c) \), which predicts the expected cumulative future rewards from state \( s_n \) by following the policy \( \pi \). Mathematically, the state value function is expressed as \cite{b19}:
\begin{equation}
V^{\pi}(s_n; \varphi_c) = \mathbb{E}_{\pi_{\varphi_a}} \left[ \sum_{\tau=0}^{\infty} \gamma^\tau \, r(s_{n+\tau}, a_{n+\tau}) \,\middle|\, s_n = s \right]
\end{equation}
where $ \gamma \in [0,1]$ is the discount factor. At the $\tau$-step horizon used by A3C, the cumulative rewards is defined by:
\begin{equation}
\Xi_n = \sum_{i=0}^{\tau-1} \gamma^i r(s_{n+i}, a_{n+i}) + \gamma^{\tau} V^{\pi} (s_{n+\tau}; \varphi_c).
\end{equation}
Moreover, the A3C uses the advantage function $\Theta(s, a) = \Xi_n - V^{\pi}(s_n; \varphi_c)$ to improve the learning stability and efficient. Specifically, the quantity $\Theta(s, a)$ is expressed as:
\begin{equation}
\begin{aligned}
\Theta (s_n , a_n ) &= \sum_{i=0}^{\tau-1} \gamma^i r(s_{n+i}, a_{n+i})  \\
& + \gamma^{\tau} V^{\pi} (s_{n+\tau}; \varphi_c) - V^{\pi} (s_n; \varphi_c).
\end{aligned}
\end{equation}
Accordingly, the actor network’s loss function that optimizes the policy performance by high-advantage actions reinforcements while regularizing for stability is expressed as: 
\begin{equation}
\begin{aligned}
L_{\pi}(\varphi_a) = \log \pi(a_n | s_n; \varphi_a) \Theta (s_n , a_n ) + \Phi \mathcal{G}(\pi(s_n; \varphi_a)).
\end{aligned}
\end{equation}
where \(\Phi\) is the hyperparameter for regularization, and  \(\mathcal{G}(\pi(s_n; \varphi_a))\) is the entropy term, which favors policy exploration. The actor's loss function $L_{\pi}(\varphi_a)$ above combines policy gradient methods with entropy regularization to balance exploitation and exploration. Its accumulated gradient \(L_{\pi}(\varphi_a)\) across threads is computed as:
\begin{equation}
\begin{aligned}
d\varphi_a &= d\varphi_a + \nabla_{\varphi'_a} \log \pi(a_n | s_n; \varphi'_a) \Theta (s_n , a_n )\\
& \hspace{1.1cm}+ \Phi \nabla_{\varphi'_a} \mathcal{G}(\pi(s_n; \varphi'_a)),
\end{aligned}
\end{equation}
where $\varphi'_a$ represents the actor network parameters specific to each thread in the asynchronous learning process. The critic's loss function that minimizes the squared advantage function is $L(\varphi_c) = \left( \Xi_n - V^{\pi}(s_n; \varphi_c) \right)^2$. In the critic gradient updates $d\varphi_c$, we combine the advantage's squared error gradient $L(\varphi_c)$ with the thread parameter $\varphi'_c$ as follows \cite{b19}:
\begin{equation}
\begin{aligned}
d\varphi_c = d\varphi_c + \frac{\partial(\Xi_n - V^{\pi}(s_n; \varphi_c))^2}{\partial \varphi'_c}
\end{aligned}
\end{equation}
To enhance training stability and convergence, we employ the Root Mean Square Propagation (RMSProp) optimizer to update the parameters with average squared gradient $\mu = \delta\mu + (1 - \delta)(\Delta\varphi)^2$ and update rule stated by:
\begin{equation}
 \quad\varphi \leftarrow \varphi - \varepsilon \frac{\Delta\varphi}{\sqrt{\mu + \alpha}} 
\end{equation}
where $\delta$ denotes momentum, $\varepsilon$ is the learning rate $\varepsilon$ and $\alpha >0$ is a small positive constant added for numerical stability.
\begin{algorithm}[!b]
\caption{Meta-A3C for Joint UAV-UGV Position and Trajectory Optimization}
\begin{algorithmic}[1]
\State \textbf{Initialization:} Global parameters $\varphi = \{\varphi_a, \varphi_c\}$; $\varphi' = \{\varphi'_a, \varphi'_c\}$; task distribution $p(\mathcal{T})$, inner/outer learning rates $\beta_{lr}^{in}$, and $\beta_{lr}^{m}$, maximum counters $N_{max}^{A3C}$ and $n_{max}^{A3C}$ and meta batch size $B$.
\State \textbf{Meta-Training:}
\For{$N = 1$ to $N_{max}^{meta}$}
    \State Sample batch of tasks $\{\mathcal{T}_i\}_{i=1}^B \sim p(\mathcal{T})$
    \For{each task $\mathcal{T}_i$}
        \State Clone parameters: $\varphi'_i \leftarrow \varphi$
        \State Initialize trajectory buffer $\mathcal{D}_i \leftarrow \emptyset$
        
        \State \textbf{Task Adaptation:}
        \For{$n = 1$ to $N_{max}^{A3C}$}
            \State Collect $\tau = (s_t,a_t,r_t,s_{t+1})$ using $\pi_{\varphi'_i}$
            \State Store $\tau$ in $\mathcal{D}_i$
            \State Compute advantages $\Theta_{\mathcal{T}_i}$ using (15)
            \State Update $\varphi'_i$ via 
            $\varphi'_i \leftarrow \varphi'_i - \beta_{lr}^{in} \nabla_{\varphi'_i} \mathcal{L}_{\mathcal{T}_i}(\varphi'_i)$
        \EndFor
        
        \State Evaluate $\mathcal{T}_i$:
        $\mathcal{L}_{\mathcal{T}_i}^{meta} \leftarrow \text{Performance}(\pi_{\varphi'_i})$
    \EndFor
    
    \State \textbf{Meta-Update:}
    \State Compute meta-gradient:
    $\nabla_\varphi^{meta} \leftarrow \nabla_\varphi \sum_i \mathcal{L}_{\mathcal{T}_i}^{meta}(\varphi'_i)$
    \State Update global parameters:
    $\varphi \leftarrow \varphi - \beta_{lr}^{m} \nabla_\varphi^{meta}$
\EndFor

\State \textbf{Online Deployment:}
\While{mission ongoing}
    \State Observe current task $\mathcal{T}_{new}$ (environment conditions)
    \State Rapid adaptation:
    $\varphi_{new} \leftarrow \varphi - \beta_{lr}^{in} \nabla_\varphi \mathcal{L}_{\mathcal{T}_{new}}(\varphi)$
    \State Execute policy $\pi_{\varphi_{new}}$ for UAV-UGV coordination:
    \State \quad • UAV 3D positioning \& obstacle avoidance
    \State \quad • UGV path planning \& terrain adaptation
    \State \quad • Cooperative target tracking
\EndWhile
\end{algorithmic}
\end{algorithm}
\subsection{Meta-Learning Integration}
We introduce a meta-learning framework that significantly enhances the adaptability of the A3C algorithm to a dynamic environment. We define task space $(\mathcal{T},p(\mathcal{T}))$ of MDP, each $ \mathcal{T}_i = \{\mathcal{S}, \mathcal{A}, \mathcal{P}_i,\mathcal{R}_i,\mathcal{\gamma} \} $ shares state and action with complete task set $ \mathcal{T}_i = \{\mathcal{T}_1, \mathcal{T}_2, \dots, \mathcal{T}_\mathcal{W}\}$. The model learns parameters $\varphi = \{\varphi_a, \varphi_c\}$ that generalize well accross $\mathcal{T}_i \sim p(\mathcal{T})$ through few gradient steps. For any task $\mathcal{T}_i$, the model uses a task-specific loss function $\mathcal{L}_{\mathcal{T}_i}(\varphi)$ to ensure rapid convergence and performance refinement. In the task loss $\mathcal{L}_{\mathcal{T}_i}(\varphi)$ below,
\begin{equation}
\mathcal{L}_{\mathcal{T}_i}(\varphi) = \mathbb{E}_{\substack{s_n \sim \rho_{\pi_\varphi} \\ a_n \sim \pi_\varphi}} \left[ -\log \pi_\varphi(a_n|s_n)  \Theta_{\mathcal{T}_i}(s_n,a_n) \right]
\end{equation}
where the advantage function $\Theta_{\mathcal{T}_i}(s_n, a_n) = \mathcal{R}_n^{\mathcal{T}_i} - V^{\pi}(s_n; \varphi_c)$ measures the $a_n$ in state $s_n$ for task $\mathcal{T}_i$  with $\mathcal{R}_n^{\mathcal{T}_i} = \sum_{i=0}^{\tau-1} \gamma^i r_{\mathcal{T}_i}(s_{n+i}, a_{n+i}) + \gamma^\tau V^{\pi}(s_{n+\tau}; \varphi_c)$. Note that $\rho_{\pi_{\varphi}}$ is the state distribution under policy $\pi_{\varphi}$. For rapid adaptation, we incorporate the Model-Agnostic Meta-Learning (MAML) approach to learn  $\varphi$ that can fine-tune to any task $\mathcal{T}_i \sim p(\mathcal{T})$. Thus, the policy gradient is expressed as:
\begin{equation}
\nabla_{\varphi} \mathcal{L}_{\mathcal{T}_i}(\varphi) = \mathbb{E}_{\mathbf{s},\mathbf{a}\sim\pi_{\varphi'}} \left[ \frac{\pi_{\varphi}(\mathbf{a}|\mathbf{s})}{\pi_{\varphi'}(\mathbf{a}|\mathbf{s})}  \Theta_{\mathcal{T}_i}(\mathbf{s},\mathbf{a})  \nabla_{\varphi} \log \pi_{\varphi}(\mathbf{a}|\mathbf{s}) \right]
\end{equation}
The term $\pi_{\varphi'}(\mathbf{a}|\mathbf{s})$ represents the policy evaluated under the old parameters $\varphi'$ for importance sampling. The new policy adapted to a specific task  $\mathcal{T}_i$ is defined as $\varphi'_{T_i} \leftarrow \varphi - \beta_{lr}^{in} \nabla_{\varphi} \mathcal{L}_{T_i}(\varphi)$, where \( \beta_{lr}^{in} \) is the inner-loop learning rate, and \( \nabla_{\varphi} \mathcal{L}_{T_i}(\varphi) \) represents the gradient of the task-specific loss function $\mathcal{L}_{T_i}(\varphi)$ with respect to the policy parameters $\varphi$. The meta-objective is to find $\varphi$ that minimizes the expected loss across all tasks after adaptation:
\begin{equation}
\min_{\varphi} \sum_{\mathcal{T}_i \sim p(\mathcal{T})} \mathcal{L}_{\mathcal{T}_i}(\varphi'_i) = \sum_{\mathcal{T}_i \sim p(\mathcal{T})} \mathcal{L}_{\mathcal{T}_i}(\varphi - \beta_{lr}^{in} \nabla_{\varphi} \mathcal{L}_{\mathcal{T}_i}(\varphi))
\end{equation}
The meta-gradient update for $\varphi$ that involves differentiation through inner loop adaptation is given by:
\begin{equation}
\varphi \leftarrow \varphi - \beta_{lr}^{m}  \nabla_{\varphi} \sum_{\mathcal{T}_i \sim p(\mathcal{T})} \mathcal{L}_{\mathcal{T}_i}(\varphi'_i)
\end{equation}
where $\beta_{lr}^{m}$ is the outer learning rate. In addition, the gradient $\nabla_{\varphi} \mathcal{L}_{\mathcal{T}_i}^{meta}(\varphi'_i)$ requires backpropagation through inner loop update
\section{Numerical Results}
In this section, we evaluate the performance of our proposed Meta-A3C framework for joint UAV-UGV positioning and trajectory optimization. The system configuration consists of $U=4$ UAVs at flight altitudes between $z_{min}=30$ m and $z_{max}=150$ m. The maximum UAV speed is 30m/s with UAV's safety distance of $10$ m. The ground network consists of $K=100$ users randomly distributed in $3000\times 3000$ m area. We want to ensure that each user receives a minimum data rate of $0.5$ Mbps. With different simulation settings, the wireless carrier frequency $f_c=2$GHz with LoS and NLoS loss coefficients $\beta^{LoS}=1$ and $\beta^{LoS}=20$, respectively. Moreover, the system consists of $4$ UGVs with constrained velocity $V^{max}=20$m/s and road speed limit of $v^{max}=15$m/s. The system considers $1W$ transmit power of $1W$ for UAVs with a noise floor of $1pW$. The actor and critic of A3C are trained with learning rates of $0,0005$ and $0,001$, respectively, and $ \gamma =0.99$. For meta learning, the configuration consists of a meta-learning rate of $0,0001$, $5$ inner steps per task, and a meta-batch size of 4 tasks per update.
\begin{figure*}[t]
  \centering
  \begin{subfigure}{0.29\textwidth}
    \centering
    \includegraphics[width=\linewidth]{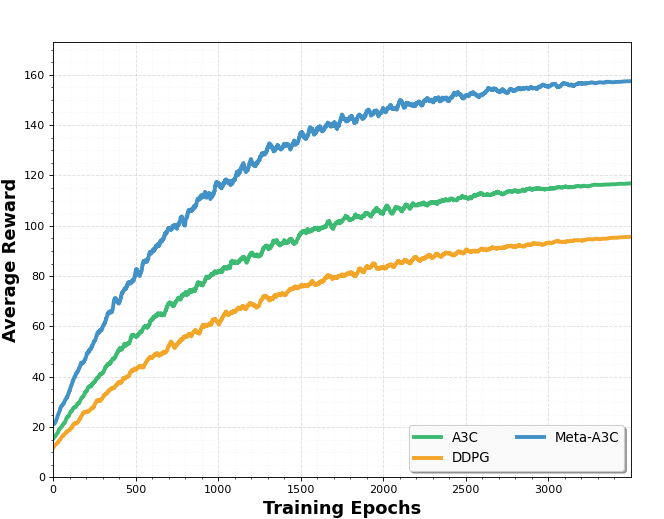}
    \caption{Convergence behavior}
    \label{fig:sub1}
  \end{subfigure}
  \hfill
  \begin{subfigure}{0.32\textwidth}
    \centering
    \includegraphics[width=\linewidth]{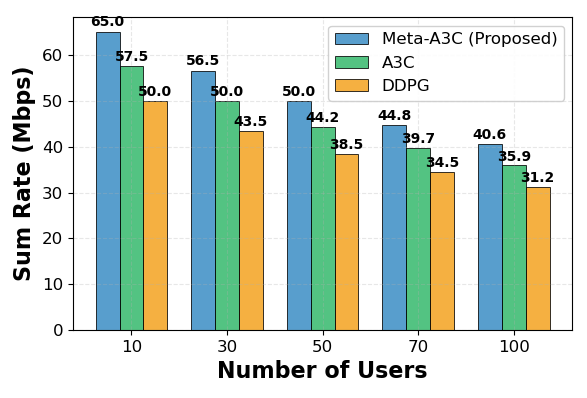}
    \caption{System performance Analysis.}
    \label{fig:sub2}
  \end{subfigure}
  \hfill
  \begin{subfigure}{0.35\textwidth}
    \centering
    \includegraphics[width=\linewidth]{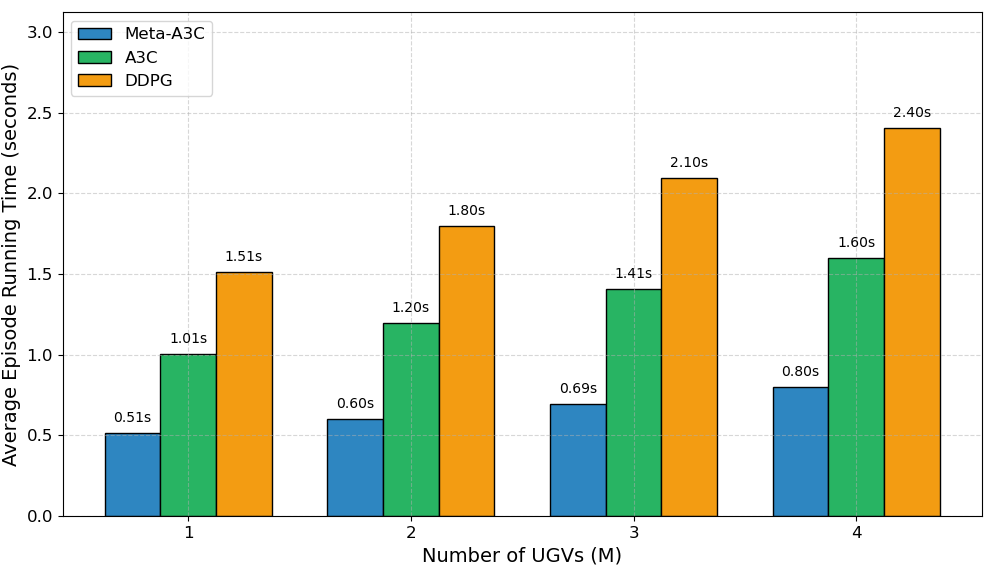}
    \caption{Complexity Analysis.}
    \label{fig:sub3}
  \end{subfigure}
  
  \caption{(a) Convergence behavior of the considered approaches over epochs, (b) sum rate performance with varying number of users for the compared approaches, (c) Complexity analysis of the proposed approach and existing RL algorithms.}
  \label{fig:three-figs-top}
  \vspace{-1.5em}
\end{figure*}

Fig. 2 (a) illustrates the convergence behavior of our proposed Meta A3C approach with other RL approaches over training epochs. As shown, Meta A3C achieves faster convergence with high reward, highlighting its adaptability in a dynamic environment. In addition, A3C outperforms DDPG while DDPG records the lowest reward among the three considered approaches.

Fig. 2(b) depicts the sum rate versus the number of users for different user configurations, with $U=4$ UAVs and $M=4$ UGVs. As demonstrated, the sum rate decreases with the increasing number of users, reflecting the need for more resources to satisfy QoS required for each user. It's worth noting that Meta-A3C outperforms A3C by approximately 13.1\% at users $K=100$ performance improvement over A3C and 30.1\% over DDPG, highlighting its effectiveness in sustaining higher throughput in post-disaster scenarios. In Fig. 2(c), we consider the average episode running time for three different algorithms to compare their complexities. The proposed Meta A3C approach shows low complexity across all UGV configurations compared to other algorithms. 
\begin{figure}[t]
  \centering
  \includegraphics[width=0.30\textwidth]{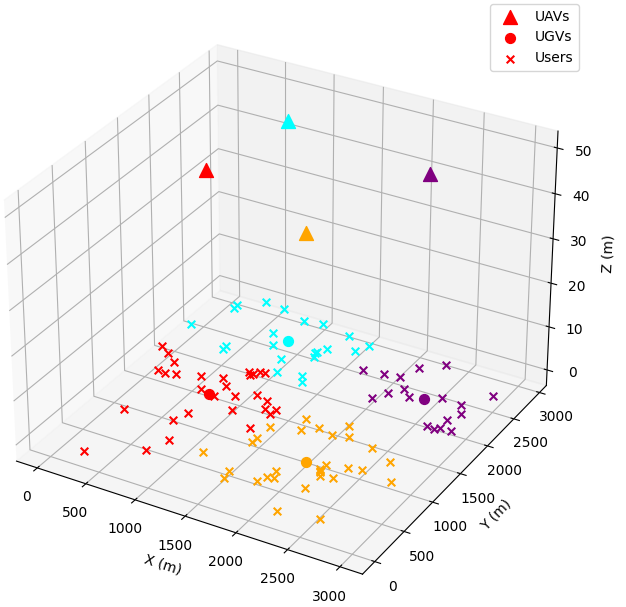}
  \caption{Optimal positioning of UAVs, UGVs, and users.}
  \label{fig:my-figure}
\end{figure}
Fig.3 illustrates the optimal 3D positioning of $U=4$ UAVs (triangular markers ) and $M=4$ UGVs (circular markers) relative to $K=100$ ground users (x markers) for one time step. Specifically, the UGVs are constrained on a road speed limit of $v^{max}=15$m/s to ensure reliable G2A links, and each UAV dynamically adjusts its altitude to meet the QoS requirement. This deployment strategy jointly optimizes the sum and enhances connectivity.  
\begin{figure}[t]
  \centering
  \includegraphics[width=0.34\textwidth]{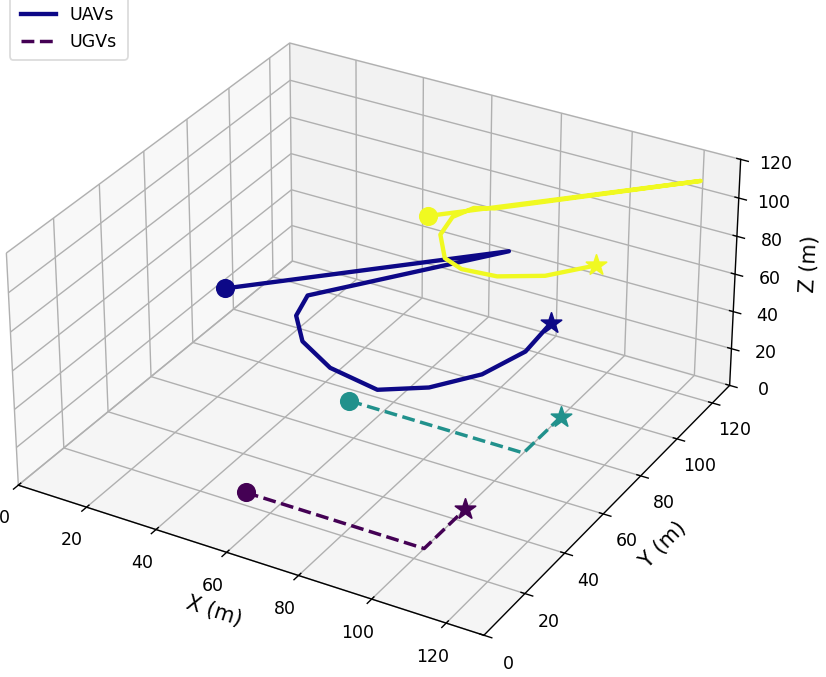}
  \caption{Optimal 3D trajectory for UAVs and UGVs.}
  \label{fig:my-figure}
\end{figure}
Fig. 4 shows the 3D trajectory of UAVs and UGVs over 25 consecutive time steps. Specifically, the UAVs' trajectories are represented by solid lines (blue and yellow), while the UGVs' trajectories are represented by dotted lines. As shown, UAV dynamically adjusts the altitude to maximize the coverage while the UGV, modeled by a road graph, optimizes its trajectory to maintain backhaul connectivity with UAVs.
\section{Conclusion}
In this work, we have investigated the joint optimization of UAV and UGV positioning and trajectory planning to maximize the sum rate while meeting the QoS requirements for users. We introduce a Meta-A3C and reformulate the non-convex optimization problem as an MDP by modeling UGV mobility with a road graph. Simulation results demonstrated that the proposed approach outperformed A3C and DDPG, achieving 13.1\% higher network throughput while meeting the QoS requirements with efficient 3D trajectory planning.
\section{ACKNOWLEDGMENT}
The work of Mehdi Sookhak was supported by the National Science Foundation (NSF) under grant number CNS-2318725.

\bibliographystyle{IEEEtran} 
\bibliography{references}

@article{b1,
  title={Enhanced emergency communication services for post-disaster rescue: Multi-IRS assisted air-ground integrated data collection},
  author={Zhou, Yi and Jin, Zhanqi and Shi, Huaguang and Shi, Lei and Lu, Ning and Dong, Mianxiong},
  journal={IEEE Transactions on Network Science and Engineering},
  year={2024},
  publisher={IEEE}
}

@article{b2,
  author    = {Munasinghe, I. and Perera, A. and Deo, R. C.},
  title     = {A comprehensive review of UAV-UGV collaboration: Advancements and challenges},
  journal   = {Journal of Sensor and Actuator Networks},
  volume    = {13},
  number    = {6},
  pages     = {81},
  year      = {2024}
}

@inproceedings{b3,
  title={Game theoretical bandwidth allocation in UAV-UGV collaborative disaster relief networks},
  author={Ying, Bincheng and Su, Zhou and Xu, Qichao and Ma, Xiandong},
  booktitle={2021 IEEE 23rd Int Conf on High Performance Computing \& Communications; 7th Int Conf on Data Science \& Systems; 19th Int Conf on Smart City; 7th Int Conf on Dependability in Sensor, Cloud \& Big Data Systems \& Application (HPCC/DSS/SmartCity/DependSys)},
  pages={1498--1504},
  year={2021},
  organization={IEEE}
}

@article{b4,
  title={Deep Reinforcement Learning Enabled Persistent Surveillance with Energy-Aware UAV-UGV Systems for Disaster Management Applications},
  author={Mondal, Md Safwan and Ramasamy, Subramanian and Bhounsule, Pranav},
  journal={arXiv preprint arXiv:2502.02666},
  year={2025}
}

@article{b5,
  author    = {Sobouti, M. J. and Adarbah, H. Y. and Alaghehband, A. and Chitsaz, H. and Mohajerzadeh, A. and Sookhak, M. and Afghah, F.},
  title     = {Efficient fuzzy-based 3-D flying base station positioning and trajectory for emergency management in 5G and beyond cellular networks},
  journal   = {IEEE Systems Journal},
  year      = {2024}
}

@article{b6,
  author    = {Rahimi, Z. and Sobouti, M. J. and Ghanbari, R. and Seno, S. A. H. and Mohajerzadeh, A. H. and Ahmadi, H. and Yanikomeroglu, H.},
  title     = {An efficient 3-D positioning approach to minimize required UAVs for IoT network coverage},
  journal   = {IEEE Internet of Things Journal},
  volume    = {9},
  number    = {1},
  pages     = {558--571},
  year      = {2021}
}

@article{b7,
  author    = {Mei, H. and Yang, K. and Liu, Q. and Wang, K.},
  title     = {3D-trajectory and phase-shift design for RIS-assisted UAV systems using deep reinforcement learning},
  journal   = {IEEE Transactions on Vehicular Technology},
  volume    = {71},
  number    = {3},
  pages     = {3020--3029},
  year      = {2022}
}

@article{b9,
  title={3D location and resource allocation optimization for UAV-enabled emergency networks under statistical QoS constraint},
  author={Niu, Haibin and Zhao, Xinyu and Li, Jing},
  journal={IEEE Access},
  volume={9},
  pages={41566--41576},
  year={2021},
  publisher={IEEE}
}

@article{b10,
  author    = {Li, Z. and Zhao, W. and Liu, C.},
  title     = {Completion time minimization for UAV-UGV-enabled data collection},
  journal   = {Sensors},
  volume    = {22},
  number    = {15},
  pages     = {5839},
  year      = {2022}
}

@article{b11,
  author    = {Ebrahimi, D. and Sharafeddine, S. and Ho, P. H. and Assi, C.},
  title     = {Autonomous UAV trajectory for localizing ground objects: A reinforcement learning approach},
  journal   = {IEEE Transactions on Mobile Computing},
  volume    = {20},
  number    = {4},
  pages     = {1312--1324},
  year      = {2020}
}

@article{b12,
  author    = {Zhang, T. and Lei, J. and Liu, Y. and Feng, C. and Nallanathan, A.},
  title     = {Trajectory optimization for UAV emergency communication with limited user equipment energy: A safe-DQN approach},
  journal   = {IEEE Transactions on Green Communications and Networking},
  volume    = {5},
  number    = {3},
  pages     = {1236--1247},
  year      = {2021}
}

@article{b13,
  author    = {Wu, H. and Lyu, F. and Zhou, C. and Chen, J. and Wang, L. and Shen, X.},
  title     = {Optimal UAV caching and trajectory in aerial-assisted vehicular networks: A learning-based approach},
  journal   = {IEEE Journal on Selected Areas in Communications},
  volume    = {38},
  number    = {12},
  pages     = {2783--2797},
  year      = {2020}
}

@article{b14,
  author    = {Wang, D. and Yang, Y.},
  title     = {Joint obstacle avoidance and 3D deployment for securing UAV-enabled cellular communications},
  journal   = {IEEE Access},
  volume    = {8},
  pages     = {67813--67821},
  year      = {2020}
}

@article{b15,
  author    = {Sivaneri, Victor O. and Gross, Jason N.},
  title     = {UGV-to-UAV cooperative ranging for robust navigation in GNSS-challenged environments},
  journal   = {Aerospace Science and Technology},
  volume    = {71},
  pages     = {245--255},
  year      = {2017}
}

@article{b16,
  author    = {Martinez-Rozas, S. and Alejo, D. and Caballero, F. and Merino, L.},
  title     = {Path and trajectory planning of a tethered UAV-UGV marsupial robotic system},
  journal   = {IEEE Robotics and Automation Letters},
  volume    = {8},
  number    = {10},
  pages     = {6475--6482},
  year      = {2023}
}

@article{b17,
  author    = {Ropero, F. and Muñoz, P. and R-Moreno, M. D.},
  title     = {TERRA: A path planning algorithm for cooperative UGV–UAV exploration},
  journal   = {Engineering Applications of Artificial Intelligence},
  volume    = {78},
  pages     = {260--272},
  year      = {2019}
}

@article{b19,
  author    = {Zhou, Y. and Jin, Z. and Shi, H. and Shi, L. and Lu, N. and Dong, M.},
  title     = {Enhanced emergency communication services for post-disaster rescue: Multi-IRS assisted air-ground integrated data collection},
  journal   = {IEEE Transactions on Network Science and Engineering},
  year      = {2024}
}

@INPROCEEDINGS{b24,
  author={Rahimi, Zahra and Ghanbari, Reza and Mohajerzadeh, Amir Hossein and Ahmadi, Hamed and Sookhak, Mehdi},
  booktitle={GLOBECOM 2022 - 2022 IEEE Global Communications Conference}, 
  title={3D UAV BS Positioning and Backhaul Management in Cellular Network Via Stochastic Optimization}, 
  year={2022},
  volume={},
  number={},
  pages={2169-2175},
  doi={10.1109/GLOBECOM48099.2022.10001472}}

@article{b25,
author = {Zhang, Wen and Pan, Chen and Liu, Tao and Zhang, Jeff (Jun) and Sookhak, Mehdi and Xie, Mimi},
title = {Intelligent Networking for Energy Harvesting Powered IoT Systems},
year = {2024},
issue_date = {March 2024},
publisher = {Association for Computing Machinery},
address = {New York, NY, USA},
volume = {20},
number = {2},
issn = {1550-4859},
url = {https://doi.org/10.1145/3638765},
doi = {10.1145/3638765},
journal = {ACM Trans. Sen. Netw.},
month = feb,
articleno = {45},
numpages = {31}
}

@INPROCEEDINGS{b27,
  title={On the Orchestration of SIM and UAV},
  author={Zarini, Hosein and An, Jiancheng and Sookhak, Mehdi and Choi, Jinho},
  booktitle={Proc. IEEE Int. Conf. Commun. (ICC)}, 
  year={2025},
  volume={},
  number={},
  pages={1-6},
  doi={10.13140/RG.2.2.18870.10560},
 note={Accepted}
}

@article{b28,
  title={Joint position and trajectory optimization of flying base station in 5G cellular networks, based on users' current and predicted location},
  author={Sookhak, Mehdi and Mohajerzadeh, Amir Hossein},
  journal={arXiv preprint arXiv:2202.03832},
  year={2022},
  pages={1-13}
}

\end{document}